\RequirePackage{fix-cm}
\documentclass[twocolumn]{svjour3}          
\smartqed
\usepackage{graphicx}
\usepackage{cite}
\usepackage{amsmath,amssymb,amsfonts}
\usepackage{algorithmic}
\usepackage{graphicx}
\usepackage{textcomp}
\usepackage{xcolor}
\usepackage{longtable,rotating}
\usepackage{threeparttable}
\usepackage{subfigure}
\usepackage{graphicx}
\usepackage{amssymb}
\usepackage{lineno}
\usepackage{multirow}
\usepackage{bm}
\usepackage{hyperref}
\usepackage{CJKutf8}
\usepackage[utf8]{inputenc}
\usepackage[ruled,linesnumbered]{algorithm2e}

\def\BibTeX{{\rm B\kern-.05em{\sc i\kern-.025em b}\kern-.08em
    T\kern-.1667em\lower.7ex\hbox{E}\kern-.125emX}}

\journalname{Neural Computing and Applications}
\begin{document}

\title{NE-LP: Normalized Entropy and Loss Prediction based Sampling for Active Learning in Chinese Word Segmentation on EHRs}

\author{Tingting Cai \and Zhiyuan Ma \and Hong Zheng \and Yangming Zhou
}
\institute{
            Corresponding author : Yangming Zhou\at
                    \email{ymzhou@ecust.edu.cn}
            \and
            Tingting Cai \at
                    School of Information Science and Engineering, East China University of Science and Technology, Shanghai 200237, China\\
                    \email{y30190775@mail.ecust.edu.cn}      
            \and
            Zhiyuan Ma \at
                    Institute of Machine Intelligence, University of Shanghai for Science and Technology, Shanghai 200093, China\\
                    \email{yuliar3514@usst.edu.cn}     
            \and
            Hong Zheng \at
                    School of Information Science and Engineering, East China University of Science and Technology, Shanghai 200237, China\\
                    \email{zhenghong@ecust.edu.cn}    
            \and
            Yangming Zhou \at
                    School of Information Science and Engineering, East China University of Science and Technology, Shanghai 200237, China\\
                    The Key Laboratory of Advanced Control and Optimization for Chemical Processes, Ministry of Education, East China University of Science and Technology, 130 Meilong Road, 200237 Shanghai, China\\
                    \email{ymzhou@ecust.edu.cn}  
}            

\date{Received: date / Accepted: date}

\maketitle

\begin{abstract}
Electronic Health Records (EHRs) in hospital information systems contain patients’ diagnosis and treatments, so EHRs are essential to clinical data mining. Of all the tasks in the mining process, Chinese Word Segmentation (CWS) is a fundamental and important one, and most state-of-the-art methods greatly rely on large-scale of manually-annotated data. Since annotation is time-consuming and expensive, efforts have been devoted to techniques, such as active learning, to locate the most informative samples for modeling. In this paper, we follow the trend and present an active learning method for CWS in EHRs. Specifically, a new sampling strategy combining Normalized Entropy with Loss Prediction (NE-LP) is proposed to select the most representative data. Meanwhile, to minimize the computational cost of learning, we propose a joint model including a word segmenter and a loss prediction model. Furthermore, to capture interactions between adjacent characters, bigram features are also applied in the joint model. To illustrate the effectiveness of NE-LP, we conducted experiments on EHRs collected from the Shuguang Hospital Affiliated to Shanghai University of Traditional Chinese Medicine. The results demonstrate that NE-LP consistently outperforms conventional uncertainty-based sampling strategies for active learning in CWS.
\end{abstract}

\section{Introduction}
\label{Sec:Introduction}

Electronic Health Records (EHRs) systematically collect patients' clinical information, such as health profiles, histories of present illness, past medical histories, examination results and treatment plans~\cite{gesulga2017barriers}. By analyzing EHRs, many useful information, closely related to patients, can be discovered~\cite{yang2014overview}. Since Chinese EHRs are recorded without explicit word delimiters~(e.g., ``\begin{CJK*}{UTF8}{gkai}糖尿病酮症酸中毒\end{CJK*}'' (diabetic ketoacidosis)), Chinese Word Segmentation (CWS) is a prerequisite for processing EHRs. Currently, state-of-the-art CWS methods usually require large amounts of manually-labeled data to reach their full potential. However, compared to general domain, CWS in medical domain is more difficult.
On one hand, EHRs involve many medical terminologies, such as ``\begin{CJK*}{UTF8}{gkai}高血压性心脏病\end{CJK*}'' (Hypertensive Heart Disease) and  ``\begin{CJK*}{UTF8}{gkai}罗氏芬\end{CJK*}'' (Rocephin), so only annotators with medical backgrounds are qualified to label EHRs. On the other hand, EHRs may involve personal privacies of patients. Therefore, they cannot be released on large scales for labeling. The above two reasons lead to the high annotation cost and insufficient training corpus for CWS in medical texts.

CWS was usually formulated as a sequence labeling task~\cite{li2012active}, which can be solved by supervised learning approaches, such as Hidden Markov Model (HMM)~\cite{eddy1998profile} and Conditional Random Field (CRF)~\cite{lafferty2001conditional}. However, these methods rely heavily on handcrafted features. To relieve the efforts of feature engineering, neural network based methods are beginning to thrive~\cite{collobert2011natural,chen2015long,liu2019neural}.
However, due to insufficient annotated training data, 
conventional models for CWS trained on open corpora often suffer from significant performance degradation when transferred to specific domains, let alone the researches are rarely dabbled in medical domain.

One solution for this obstacle is to use active learning, where only a small scale of samples are selected and labeled in an active manner. 
Active learning methods are favored by the researchers in many Natural Language Processing (NLP) tasks, such as text classification~\cite{tang2018combining} and Named Entity Recognition (NER)~\cite{huang2018low}. 
However, only a handful of works are conducted on CWS~\cite{li2012active}, and few focuses on medical domain. 

Given the aforementioned challenges and current researches, we propose a word segmentation method based on active learning.
To select the most informative data, we incorporate a sampling strategy called NE-LP, which consists of Normalized Entropy (NE) and Loss Prediction (LP).
Specifically, we leverage the normalized entropy of class posterior possibilities from Bi-directional Long-Short Term Memory and Conditional Random Field (BiLSTM-CRF) based word segmenter to define uncertainty. Then, we attach a ``loss prediction model" based on self-attention~\cite{vaswani2017attention} to the word segmenter and it aims to predict the loss of input data.
The final decision on the selection of labeling samples is made by calculating the sum of normalized token entropy and losses according to a certain weight. 
Besides, to capture coherence over characters, we additionally add n-gram features to the input of the joint model and experimental results show that for specific texts, such as our medical texts, bigram performs best.

To sum up, the main contributions of our work are summarized as follows:
\begin{itemize}
\item We propose a novel word segmentation method incorporating active learning and hybrid features. The former lightens the burden of labeling large amounts of data, and the latter combines bigram features with character embeddngs to achieve better representations of the coherence between adjacent characters. 
\item To improve the performance of active learning, we propose a simple, yet effective sampling strategy called NE-LP, which is based on a joint model including a word segmenter and a loss prediction model. Instead of solely relying on the uncertainty of classifying boundary to choose the most representative samples for labeling, our proposed method utilizes normalized token entropy to estimate the uncertainty from outputs of the word segmenter at statistical level, moreover, we also employ self-attention as a loss prediction model to simulate human understanding of words from the deep learning level.
\item Instead of evaluating the performance in simulated data, we use cardiovascular diseases data collected from the Shuguang Hospital Affiliated to Shanghai University of Traditional Chinese Medicine to illustrate the improvements of the proposed method. Experimental results show that NE-LP is superior to  mainstream uncertainty-based sampling strategies in $F_{1}$-score.
\end{itemize}

The rest of this paper is organized as follows. Section \ref{Sec:Related Work} briefly reviews the related work on CWS and active learning. Section \ref{Sec:Our Method} details the proposed method for CWS, followed by experimental evaluations as Section \ref{Sec:Analysis and Discussion}. In the end, the conclusions and potential research directions are summarized as Section \ref{Sec:Conclusion}.

\section{Related Work}
\label{Sec:Related Work}

\subsection{Chinese Word Segmentation}
Due to the practical significance~\cite{liu2019neural}, CWS has attracted considerable research efforts, and a great number of solution methods have been proposed in the literature in past decades~\cite{xue2003chinese,zheng2013deep,shao2019domain}. Generally, all the existing approaches fall into two categories: statistical machine learning and deep learning~\cite{li2012active}.

\textbf{Statistical Machine Learning Methods}.  Initially, statistical machine learning methods were widely-used in CWS. Xue and Shen~\cite{xue2003chinese} employed a maximum entropy tagger to automatically assign Chinese characters. Zhao et al.~\cite{zhao2006effective} used CRF for tag decoding and considered both feature template selection and tag set selection. However, these methods greatly rely on manual feature engineering~\cite{peng2004chinese}, while handcrafted features are difficult to design, and the sizes of these features are too large for practical use~\cite{chen2015long}. In such a case, deep learning methods have been increasingly employed for the ability to minimize the efforts in feature engineering.

\textbf{Deep Learning Methods}. Recently, researchers tended to apply various neural networks for CWS and achieved remarkable performance.
To name a few, Zheng et al.~\cite{zheng2013deep} used deep layers of neural networks to learn feature representations of characters.
Chen et al.~\cite{chen2015long} adopted LSTM to capture the previous important information.
Wang and Xu~\cite{wang2017convolutional} proposed a Convolutional Neural Network (CNN) to capture rich n-gram features without any feature engineering.
Gan and Zhang~\cite{gan2019investigating} investigated self-attention for CWS and observed that self-attention gives highly competitive results.
Jiang and Tang~\cite{jiang2019seq} proposed a sequence-to-sequence transformer model to avoid overfitting and capture character information at the distant site of a sentence.
La Su and Liu~\cite{la2020research} presented a hybrid word segmentation algorithm based on Bi-directional Gated Recurrent Unit (BiGRU) and CRF to learn the semantic features of the corpus.
Ma et al.~\cite{ma2018state} found that BiLSTM can achieve better results on many of the popular CWS datasets as compared
to models based on more complex neural network architectures. Therefore, in this paper, we adopt BiLSTM-CRF as our base word segmenter due to its simple architecture, yet remarkable performance.

\textbf{Open-source CWS Tools}. In recent years, more and more open-source CWS tools are emerging, such as Jieba and PyHanLP. These tools are widely-used due to convenience and great performance for CWS in general fields. However, terminologies and uncommon words in medical fields would lead to the unsatisfactory performance of segmentation results. We experimentally compare seven well-known open-source CWS tools on EHRs. As shown in Table \ref{tab:4}, we find that since these open-source tools are trained from general domain corpora, the results are not ideal enough to cater to the needs of subsequent NLP tasks when applied to medical fields.

\textbf{Domain-Specific CWS Methods}. Currently, a handful of domain-specific CWS approaches have been studied, but they focused on decentralized domains. In the metallurgical field, Shao et al.~\cite{shao2019domain} proposed a domain-specific CWS method based on BiLSTM model. In the medical field, Xing et al.~\cite{xing2018adaptive} proposed an adaptive multi-task transfer learning framework to fully leverage domain-invariant knowledge from high resource domain to medical domain. Meanwhile, transfer learning still greatly focuses on the corpora in general domain. When it comes to the specific field, large amounts of manually-annotated data are necessary. Active learning can solve this problem to a certain extent, where a model asks human to annotate data that it is uncertain of~\cite{yoo2019learning}. However, due to the challenges faced by performing active learning on CWS, only a few studies have been conducted. On judgements, Yan et al.~\cite{yan2017active} adopted the local annotation strategy, which selects substrings around the informative characters in active learning. However, their method still stays at the statistical level. Therefore, compared to the above method, we intend to utilize a new active learning approach for CWS in medical text, which combines normalized entropy with loss prediction to effectively reduce annotation cost.

\subsection{Active Learning}
Active learning~\cite{angluin1988queries} mainly aims to ease data collection process by automatically deciding which instances should be labeled by annotators, thus saving the cost of annotation~\cite{Hu2018Active}. In active learning, the sampling strategy plays a key role. Over the past few years, the rapid development of active learning has resulted in various sampling strategies, such as uncertainty sampling~\cite{lewis1994sequential}, query-by-committee~\cite{gilad2006query} and information gain~\cite{houlsby2011bayesian}.

Currently, in sequence labeling tasks, uncertainty-based method has attracted considerable attention since it performs well and saves much time in most cases~\cite{liu2020ltp}. Traditional uncertainty-based sampling strategies mainly include least confidence, maximum token entropy and minimum token margin.

\textbf{Least Confidence (LC)}. The LC strategy selects the samples whose most likely sequence tags that the model is least confident of. Despite its simplicity, this approach has been proven effective in various tasks~\cite{yoo2019learning}.

\begin{equation}
\mathcal{S}^{L C}({x})=1-p\left({y}^{*} | {x} \right)
\end{equation}
where $x$ is the instance to be predicted and $y^{*}$ represents the most likely tag sequence of $x$.

\textbf{Maximum Token Entropy (MTE)}. The MTE strategy evaluates the uncertainty of a token by entropy. The closer the distribution of marginal probability to uniform, the larger the entropy:

\begin{equation}
\mathcal{S}^{M T E}({x})= - \sum_{i=1}^{N}  p\left({y}^{*}| {x}\right) \cdot \log p\left({y}^{*}| {x}\right)
\end{equation}
where $N$ represents the number of classes.

\textbf{Minimum Token Margin (MTM)}. To measure the informativeness, MTM considers the first and second most likely assignments and subtracts the highest probability by the lowest one~\cite{marcheggiani2014experimental}:

\begin{equation}
\mathcal{S}^{M T M}({x}) = \max  p\left( {y}^{*} | {x} \right)-\max {\prime} p\left({y}^{*} | {x} \right)
\end{equation}
where $max^{\prime}$ means the second maximum probability.

However, in some complicated tasks, such as CWS and NER, only considering the uncertainty of data is obviously not enough. Therefore, we further take loss values into account and pick up samples from two perspectives including both uncertainty and loss.

\section{Joint Model Incorporated Active Learning framework for Chinese Word Segmentation}
\label{Sec:Our Method}

\subsection{Overview}
\label{SubSec:Overview}

\begin{figure}
\includegraphics[width=0.5\textwidth]{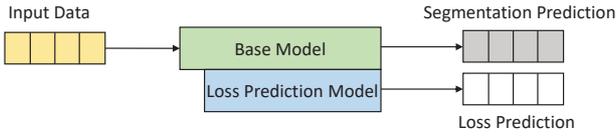}
\caption{The overall architecture of the joint model, where the loss prediction model is attached to the base model.}
\label{Fig:F1}
\end{figure}

Active learning algorithm is generally composed of two parts: a learning engine and a selection engine. The learning engine is essentially a segmenter, which is mainly utilized for training in sequence labeling problems. The selection engine picks up unlabeled samples based on preset sampling strategy and submits these samples for human annotation. Then, we incorporate them into the training set after experts complete the annotation, thus continuously improving the $F_{1}$-score of the segmenter with the increasing of the training set size~\cite{song-etal-2010-active}. In this paper, we propose a joint model as a selection engine. Fig.~\ref{Fig:F1} shows the overall architecture of the joint model, where the loss prediction model predicts the loss value from input data. Moreover, the loss prediction model is (i) attached to the base model, and (ii) jointly learned with the base model. Here, the base model is employed as a learning engine.

Algorithm~\ref{alg:1} demonstrates the procedure of CWS based on active learning with the sampling strategy of NE-LP. First, with training set, we train a joint model including a segmenter and a loss prediction model. Later, the joint model selects $n$-highest ranking samples based on NE-LP strategy, which are expected to improve the performance of the segmenter to the largest extent. Afterwards, medical experts annotate these instances manually. Finally, these annotated instances are incorporated into the training set, and we use the new training set to train the joint model. The above steps iterate until the desired $F_{1}$-score is achieved or the number of iterations has reached a predefined threshold.

\begin{algorithm}[!ht]
\caption{NE-LP based Active Learning for Chinese Word Segmentation}
\label{alg:1}
\KwIn{ labeled data $L$, unlabeled data $U$, the number of iterations $M$, the number of samples selected per iteration $n$, partitioning function $Split$, size $\tau$}
\KwOut{a word segmentation model $f^*$ with the smallest testing set loss $l_{min}$}
\Begin{
\textbf{Initialize:} 	$Training_\tau,Testing_\tau\leftarrow Split(L,\tau)$\\
	\qquad\qquad train a joint model with a word segmenter $f_\tau$ and a loss prediction model $t_\tau$\\
	\qquad\qquad estimate the testing set loss $l_\tau$ on $f_\tau$\\
	\qquad\qquad label $U$ by $f_\tau$\\
\For{$i=1$ to $M$}{
    \For{$Sample\in U$}{
    compute $Uncertainty_{Sample}$ from the output of $f$ and predict $Loss_{Sample}$ by $t$\\
    calculate the sum of $Uncertainty_{Sample}$ and $Loss_{Sample}$ according to a certain weight\\
    }
    select $n$-highest ranking samples $R$\\
    relabel $R$ by annotators\\
    form a new labeled dataset $Training_R\leftarrow Training_\tau \bigcup \{R\}$\\
    form a new unlabeled dataset $U_R\leftarrow U_\tau\backslash \{R\}$\\
    train a joint model with  $f_R$ and $t_R$\\
    estimate the new testing loss $l_R$ on $f_R$\\
    compute the loss reduction $\delta_R \leftarrow l_R- l_\tau$\\
    \If{$\delta _R < 0$}{
    $l_{min} \leftarrow l_R$\\
    }
    \Else{$l_{min} \leftarrow l_\tau$\\}
}
$f^*\leftarrow$ $f$ with the smallest testing set loss $l_{min}$
}
\Return $f^*$
\end{algorithm}

\begin{figure*}
\includegraphics[width=1\textwidth]{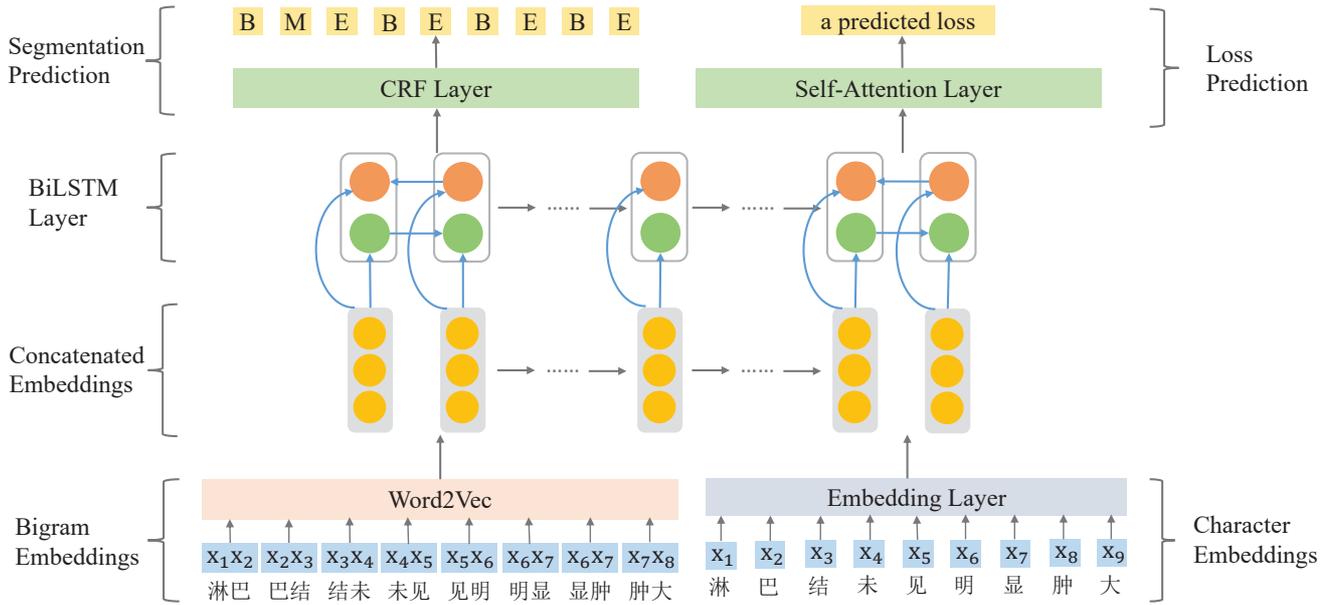}
\caption{The detailed architecture of the joint model, where BiLSTM-CRF is employed as a word segmenter and BiLSTM-Self-Attention is a loss prediction model. The loss prediction model shares BiLSTM layer parameters with word segmenter to learn feature representations better for loss prediction.}
\label{Fig:F2}
\end{figure*}

Fig.~\ref{Fig:F2} demonstrates the detailed architecture of the joint model. First, we pre-process EHRs at the character-level, separating each character of raw EHRs. For instance, given a sentence $L=\left[C_0C_1C_2 \ldots C_{n-1}C_n \right]$, where $C_i$ represents the $i$-th character, the separated form is $L_s=\left[C_0,C_1,C_2,\ldots,C_{n-1},C_n\right]$, and we obtain the character embeddings by converting character indexes into fixed dimensional dense vectors. Afterwards, to capture interactions between adjacent characters, bigram embeddings are utilized to feature the coherence over characters. We construct the bigram feature for each character by concatenating it with the previous character, i.e., $B = \left[x_0x_1, x_1x_2,\ldots, x_{t-1}x_t\right]$. We employ Word2Vec~\cite{goldberg2014word2vec} to train bigram features to get bigram embedding vectors. Then, we concatenate the character embeddings and bigram embeddings as the input of BiLSTM layer. Finally, CRF layer makes positional tagging decisions over individual characters, and self-attention layer learns to simulate the loss defined in the base model.

\subsection{BiLSTM-CRF based Word Segmenter}
\label{SubSec:Bi-LSTM-CRF based Word Segmenter}

CWS can be formalized as a sequence labeling problem with character position tags, which are (`B', `M', `E', `S'), so we convert the labeled data into the `BMES' format, in which each character in the sequence is assigned with a label as follows: B=beginning of a word, M=middle of a word, E=end of a word and S=single word. For example, a Chinese segmented sentence \begin{CJK}{UTF8}{gkai}``病人/长期/于/我院/心血管科/住院/治疗/。/"\end{CJK}~(The patient was hospitalized for a long time in the cardiovascular department of our hospital.) can be labeled as `BEBESBEBMMEBEBES'. In this paper, we use BiLSTM-CRF as the base model for CWS, which is widely-used in sequence labeling.
\subsubsection{BiLSTM Layer}
LSTM is mainly an optimization for traditional Recurrent Neural Network (RNN). RNN is widely used to deal with time-series prediction problems. The result of its current hidden layer is determined by the input of the current layer and the output of the previous hidden layer~\cite{lei2018effective}. Therefore, RNN can remember historical results. However, traditional RNN has vanishing gradient and exploding gradient problems when training long sequences~\cite{bengio1994learning}, and LSTM can effectively solve these problems by adding a gated mechanism to RNN. Formally, the LSTM unit performs the following operations at time step $t$:
\begin{equation}
f_{t}=\sigma_{g}\left(W_{f} x_{t}+U_{f} h_{t-1}+b_{f}\right)
\end{equation}
\begin{equation}
i_{t}=\sigma_{g}\left(W_{i} x_{t}+U_{i} h_{t-1}+b_{i}\right)
\end{equation}
\begin{equation}
o_{t}=\sigma_{g}\left(W_{o} x_{t}+U_{o} h_{t-1}+b_{o}\right)
\end{equation}
\begin{equation}
c_{t}=c_{t-1} \odot f_{t}+i_{t} \odot \sigma_{c}\left(W_{c} x_{t}+U_{c} h_{t-1}+b_{c}\right)
\end{equation}
\begin{equation}
h_{t}=\sigma_{h}\left(c_{t}\right) \odot o_{t}
\end{equation}
where $x_t$, $c_{t-1}$, $h_{(t-1)}$ are the inputs of LSTM, all $W_*$ and $U_*$ are a set of parameter matrices, and $b_*$ is a set of bias parameter matrices. $\odot$ and $\sigma$ operation represent matrix element-wise multiplication and sigmoid function, respectively. In the LSTM unit, there are two hidden layers ($h_t$, $c_t$), where $c_t$ is the internal memory cell for dealing with vanishing gradient, while $h_t$ is the main output of the LSTM unit for complex operations in subsequent layers.

Obviously, the hidden state $h_t$ of the current LSTM unit only relies on the previous hidden state
$h_{t-1}$, while ignoring the next hidden state $h_{t+1}$. However, future information from the backward direction is also useful to CWS~\cite{tang2020recognizing}. BiLSTM, which consists of two LSTMs, i.e., forward LSTM and backward LSTM, can capture and merge features both from the forward and backward direction of a sequence. Therefore, BiLSTM can understand the syntactic and semantic context from a deeper perspective than LSTM. Assume that the output sequence of hidden states of the forward and backward LSTM  are $\overrightarrow{h_{t}}$ and  $\overleftarrow{h_{t}}$, respectively, the context vector can be denoted by concatenating the two hidden vectors as $h_{t}$=[$\overrightarrow{h_{t}}$ ; $\overleftarrow{h_{t}}$].

\subsubsection{CRF Layer}
For CWS, it is necessary to consider the
dependencies of adjacent tags. For example, a B (Begin) tag should be followed by an M (Middle) tag or an E (End) tag, and cannot be followed by an S (Single) tag. Given the observed sequence, CRF has a single exponential model for the joint probability of the entire sequence of labels, so it can solve the label bias problem effectively, which motivates us to use CRF to model the tag sequence jointly, not independently~\cite{wang2019incorporating}.

$A$ is an important parameter in CRF called a transfer matrix, which can be set manually or learned by model. $A_{y_{i},y_{i+1}}$ denotes the transition probability from label $y_{i}$ to $y_{i+1}$. $y^{*}$ represents the most likely tag sequence of x and it can be formalized as:

\begin{equation}
{y}^{*}=\arg \max _{{y}} p({y}|{x};A)
\end{equation}

\subsection{Self-Attention based Loss Prediction Model}
\label{SubSec:Information Entropy Based Scoring Model}

To select the most appropriate sentences in a large number of unlabeled corpora, we attach a self-attention based loss prediction model to the base word segmenter, which is inspired by~\cite{yoo2019learning}. The word segmenter is learned by minimizing the losses. If we can predict the losses of input data, it is intuitive to choose samples with high losses, which tend to be more beneficial to current segmenter improvement.

\subsubsection{Self-Attention Layer}
The attention mechanism was first proposed in the field of computer vision, and it is widely used in NLP tasks in recent years, which imitates human beings to address problems focusing on important information from big data~\cite{sun2020attention}. Attention mainly aims to map a query to a series of key-value pairs~\cite{cheng2020sentiment}. Formally, attention performs the following three operations:

\begin{enumerate}

\item Calculate the similarity between query and each
key to get the weight coefficient of the value
corresponding to each key, and then scale the dot products by $\frac{1}{\sqrt{d_{k}}}$:

\begin{equation}
f\left(Q, K_{i}\right)=\frac{Q^{T} K_{i}}{\sqrt{d_{k}}}
\end{equation}
where $d_{k}$ denotes the dimension of key and value.

\item Normalize the weight coefficient by softmax function:

\begin{equation}
a_{i}=\operatorname{softmax}\left(f\left(Q, K_{i}\right)\right)=\frac{\exp \left(f\left(Q, K_{i}\right)\right)}{\sum_{j=1}^{Len} \exp \left(f\left(Q, K_{j}\right)\right)}
\end{equation}

\item The final attention is a weighted sum of weight coefficients and values:

\begin{equation}{Attention}(Q, K, V)=\sum_{i=1}^{Len} a_{i} * V_{i}\end{equation}
where $Len$ is the length of the input sequence.
\end{enumerate}

 Self-attention mechanism is a special form of attention, where $Q$, $K$ and $V$ have the same value, i.e., each token in the sequence will be calculated attention with other remaining tokens. Self-attention can learn the internal structure of the sequence and it is more sensitive to the difference between input and output, so we use self-attention to learn the loss of word segmenter and we define that a sequence with higher self-attention score has higher loss.

\begin{figure}
\includegraphics[width=0.5\textwidth]{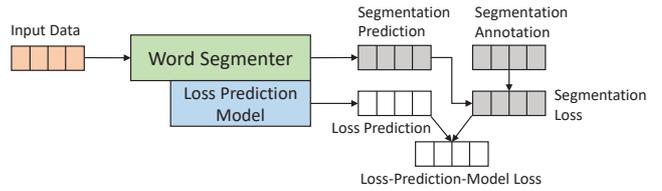}
\caption{The method for training loss prediction model. Given an input, the word segmenter and loss prediction model output a segmentation prediction and a predicted loss, respectively. Next, a segmentation loss can be computed by the segmentation prediction and annotation. Then, the segmentation loss is regarded as a ground-truth loss for the loss prediction model, and is used to compute the loss-prediction-model loss.}
\label{Fig:loss}
\end{figure}

\subsubsection{Loss Learning }
Fig. \ref{Fig:loss} shows a detailed description of how to train the loss prediction model. Given the input data $x$, the segmentation prediction can be obtained through the word segmenter: $s_{pre}$ = $Seg$($x$). Similarly, we can get the loss prediction through the loss prediction model: $loss_{pre}$ = $Loss$($x$). Next, the segmentation loss can be computed as: $loss_{Seg}$ = $L_{Seg}$($s_{pre}$, $s_{true}$), where $s_{true}$ represents the true annotation of $x$. Then, $loss_{Seg}$ is regarded as a ground-truth target for loss prediction model, so we can compute the loss of loss prediction model as $loss_{Loss}$ = $L_{Loss}$($loss_{pre}$, $loss_{Seg}$). The final loss function of the joint model is defined as:

\begin{equation}
L_ {joint } = L_ {Seg }(s_{pre}, s_{true})+ \lambda L_{Loss }(loss_{pre}, loss_{Seg})
\end{equation}
where $\lambda$ represents the weight coefficient. In the following part, we empirically set $\lambda$ to 1.

When training the loss prediction model, we seek to minimize the segmentation loss and the predicted loss:

\begin{equation}
L_ {Loss }=\frac{1}{n}\sum_{i=1}^{n}[loss_{pre} - loss_{Seg}]^2
\end{equation}
where $n$ means the number of samples.

\subsection{NE-LP Sampling Strategy}
\label{SubSec:NE-LP sampling strategy}

To judge whether the samples are effective to improve the model performance, we combine the normalized entropy of segmentation prediction with loss prediction. The former measures the uncertainty, which can be computed as Equation (15), while the latter takes segmentation loss into consideration.

\begin{equation}
{Uncertainty({x})}=\frac{\sum_{i=1}^{N} p_{Seg}(x) \log p_{Seg}(x)}{\log \frac{1}{N} \sqrt{Len}}
\end{equation}
where $p_{Seg}$ represents the output probability of word segmenter, and $N$ denotes the number of labeled classes. To ensure that the normalized entropy and loss are in the same order of magnitude, we scale the normalized entropy by $\frac{1}{\sqrt{Len}}$, where $Len$ is the length of the input sequence.

For CWS, we hypothesize that if a sample has both high uncertainty and high loss, it is probably informative to the current word segmenter, and we verify this assumption in our experiments. Therefore, the final sampling strategy NE-LP can be formalized as:

\begin{equation}
\mathcal{S}^{NE-LP}({x})= \alpha \frac{\sum_{i=1}^{N} p_{Seg}(x) \log p_{Seg}(x)}{\log \frac{1}{N} \sqrt{Len}} + \beta {Loss}({x})
\end{equation}
where $\alpha$ and $\beta$ are the weight coefficients of normalized entropy and loss prediction, respectively.

\section{Experiments \& Analysis}
\label{Sec:Analysis and Discussion}
\subsection{Datasets}

We collect 204 EHRs with cardiovascular diseases from the Shuguang Hospital Affiliated to Shanghai University of Traditional Chinese Medicine and each contains 27 types of records. We choose 4 different types with a total of 3868 records from them, which are hospital records, medical records, ward round records and discharge records. The detailed information of EHRs are listed in Table~\ref{tab:1}.

\newcommand{\tabincell}[2]{\begin{tabular}{@{}#1@{}}#2\end{tabular}}
\begin{table}
\caption{Detailed Information of EHRs.}
\label{tab:1}
\begin{tabular}{lll}
\hline\noalign{\smallskip}
Types      & Counts & Contents \\
\noalign{\smallskip}\hline\noalign{\smallskip}
Hospital records     & 957       & \tabincell{l}{Admission date,\\ history of present illness.}  \\
Medical records      & 992       & \tabincell{l}{Chief complaints,\\ physical examination.}   \\
Ward round records & 952      & \tabincell{l}{General, heart rate,\\ laboratory findings.}     \\
Discharged records & 967      & \tabincell{l}{Treatment plans,\\
date of discharge.}\\
\noalign{\smallskip}\hline
\end{tabular}
\end{table}

We divide 3868 records including 27442 sentences into training set, testing set and validation set with the ratio of 6:2:2. Then, we randomly select 4950 sentences from training set as initial labeled set, and the remaining 11525 sentences as unlabeled set, i.e., we obtain the initial labeled set and unlabeled set by splitting the training set according to the ratio of 3:7. Statistics of datasets are listed in Table~\ref{tab:2}.

\begin{table}
\caption{Statistics of Datasets.}
\label{tab:2}
\begin{tabular}{llll}
\hline\noalign{\smallskip}
Datasets      & Sentences & Words & Characters \\
\noalign{\smallskip}\hline\noalign{\smallskip}
Training set     & 16465      & 400878 & 706362      \\
Initial labeled set  & 4950      & 120699 & 212598      \\
Unlabeled set  & 11525      & 280179 & 493764      \\
Testing set      & 5489       & 131624  & 233759      \\
Validation set      & 5489      & 135406     & 238954     \\ \noalign{\smallskip}\hline
\end{tabular}
\end{table}

\subsection{Parameter Settings}

Hyper-parameter configuration may have a great impact on the performance of neural network. The hyper-parameter configurations of our method are listed in Table~\ref{tab:3}.

\begin{table}
\caption{Hyper-parameter Setting.}
\label{tab:3}
\begin{tabular}{ll}
\hline\noalign{\smallskip}
Hyper-parameters    & Setting \\
\noalign{\smallskip}\hline\noalign{\smallskip}
Maximum sequence length & $len$ = 200   \\
Character embedding dimension & $d_{cha}$ = 128 \\
Bigram embedding dimension & $d_{big}$ = 128   \\
Concatenated embedding dimension & $d_{con}$ = 256\\
BiLSTM hidden unit number & $n_{hid}$ = 512\\
Dropout rate & $rate$ = 0.2   \\
\noalign{\smallskip}\hline
\end{tabular}
\end{table}

We initialize bigram embeddings via Word2Vec on the whole datasets. The dimension of character embeddings is set as same as the bigram embeddings. Then, we concatenate two embeddings with the dimsension of 256 as the input of BiLSTM layer. BiLSTM hidden unit number is twice the dimension of concatenated embeddings. Dropout~\cite{srivastava2014dropout} is applied to the outputs of BiLSTM layer in order to prevent our model from overfitting.

In active learning, we fix the number of iterations at 10 since each sampling strategy does not improve obviously after 10 iterations. At each iteration, we select 1000 sentences from unlabeled data for joint model to learn.

\subsection{Experimental Results}

\subsubsection{Comparisons between Different Open-source CWS Tools}

We select seven widely-used and mainstream open-source CWS tools from the Internet, which are SnowNLP\footnote{https://github.com/isnowfy/snownlp}, PyHanLP\footnote{https://github.com/hankcs/pyhanlp}, Jieba\footnote{https://github.com/fxsjy/jieba}, THULAC\footnote{https://github.com/thunlp/THULAC-Python}, PyNLPIR\footnote{https://github.com/tsroten/pynlpir}, FoolNLTK\footnote{https://github.com/rockyzhengwu/FoolNLTK} and pkuseg\footnote{https://github.com/lancopku/pkuseg-python}. We evaluate them on our datasets with a total of 27443 sentences.

\begin{table}
\caption{Experimental Results of Different Open-source CWS Tools.}
\label{tab:4}
\begin{tabular}{llll}
\hline\noalign{\smallskip}
CWS tools & Precision      & Recall      & $F_{1}$-score  \\
\noalign{\smallskip}\hline\noalign{\smallskip}
SnowNLP  & 59.4   & 56.68  & 58.04          \\
PyHanLP  & 65.01  & 70.89  & 67.82         \\
Jieba    & 70.36  & 71.48   & 70.91     \\
THULAC   & 68.67  & \textbf{77.36}  & 72.76          \\
PyNLPIR  & 69.14 & 76.89     & 72.81 \\
FoolNLTK   & 72.85   & 76.98  & 74.86         \\
pkuseg  & \textbf{78.93}    & 75.86    & \textbf{77.37}   \\
\noalign{\smallskip}\hline
\end{tabular}
\end{table}

As shown in Table~\ref{tab:4}, we find that pkuseg performs the best with the $F_{1}$-score of 77.37\% while SnowNLP shows the lowest of 58.04\%, and THULAC has the highest recall of 77.36\%. However, since these open-source tools are trained by general domain corpora, when applied to specific fields, such as medical domain, the results are still not satisfactory. Therefore, we need to train a new segmenter on medical texts.

\subsubsection{Comparisons between Different Models for CWS}

To select a base word segmenter that is most suitable for medical texts, we compare different types of models including both statistical machine learning and deep learning. These models are trained on the whole training set with 20 epoches. The results are listed in TABLE~\ref{tab:model}.

All deep neural networks obtain higher $F_{1}$-score than statistical machine learning model CRF by the margins between 2.35\% and 11.84\% since neural networks can effectively model feature representations.

We further observe that self-attention-CRF shows relatively low $F_{1}$-score of 86.48\% since only a single self-attention layer cannot extract useful feature representations. Thus, to capture more features, we employ Transformer-CRF, i.e., we use the encoder part of the model proposed by~\cite{vaswani2017attention} as the feature extractor, which is composed of a multi-head attention sub-layer and a position-wise fully connected feed-forward network. Results show that Transformer-CRF has an $F_{1}$-score of 91.25\%, which is a 4.77\% improvement compared to self-attention-CRF.

Among bi-directional RNNs, BiLSTM-CRF shows a highest $F_{1}$-score of 95.89\%, while BiRNN-CRF and BiGRU-CRF achieve 95.30\% and 95.71\%, respectively. BiLSTM and BiGRU are optimizations for BiRNN since they introduce gated mechanism to solve the problem of long-distance dependencies, where BiLSTM contains three gates, which are forget, input and output gates, while BiGRU has two gates, which are reset and update gates.

Furthermore, we notice that BiLSTM-CRF outperforms LSTM-CRF by the margins of 2.78\%, which shows that BiLSTM can understand the syntactic and semantic contexts better than LSTM. Compared to CNN-CRF, the $F_{1}$-score of BiLSTM-CRF improves by 1.74\%. However, CNN is able to extract more local features, while BiLSTM may ignore some key local contexts important for CWS when modeling the whole sentence. Therefore, when combining BiLSTM and CNN as feature extractor, the $F_{1}$-score reaches the peak of 95.97\%, which outperforms BiLSTM-CRF by a small margin of 0.08\%.

Given the above experimental results, considering the computational cost, complextity of model architecture and final results, we adopt BiLSTM-CRF as our base segmenter since the performance does not improve greatly when incorporating CNN, but it costs more time due to a more complex architecture.

\begin{table}
\caption{Experimental Results of Different Models for CWS.}
\label{tab:model}
\begin{tabular}{llll}
\hline\noalign{\smallskip}
Model        & Precision      & Recall  & $F_{1}$-score             \\
\noalign{\smallskip}\hline\noalign{\smallskip}
CRF      & 83.39          & 84.88          & 84.13          \\
Self-Attention-CRF & 85.74   & 87.23  & 86.48 \\
Transformer-CRF  & 90.72   & 91.78  & 91.25 \\
LSTM-CRF & 92.76   & 93.46  & 93.11 \\
CNN-CRF   & 93.73   & 94.58   & 94.15          \\
BiRNN-CRF   & 94.90  & 95.71   & 95.30        \\
BiGRU-CRF   & 95.36  & 96.06 & 95.71          \\
BiLSTM-CRF   & \textbf{95.81}  &95.97   & 95.89 \\
CNN-BiLSTM-CRF   & 95.77  & \textbf{96.18}   & \textbf{95.97}  \\
\noalign{\smallskip}\hline
\end{tabular}
\end{table}

\subsubsection{Effectiveness of N-gram Features in BiLSTM-CRF based Word Segmenter}

To investigate the effectiveness of n-gram features in BiLSTM-CRF based word segmenter, we also compare different n-gram features on EHRs. The results are shown in Table~\ref{tab:5}.

By using additional n-gram features in BiLSTM-CRF based word segmenter, there is an obvious improvement of $F_{1}$-score, where bigram features achieve 97.70\% while trigram and four-gram reach 97.32\% and 96.71\%, respectively. Specifically, bigram, trigram and four-gram features outperform character-only features by margins of 1.81\%, 1.43\% and 0.82\%, which indicates that n-gram features can effectively capture the semantic coherence between characters.

Furthermore, we explore the reason why bigram features perform better than trigram and four-gram. We analyze the number of words consisting of 2, 3 and 4 characters in our datasets. As shown in Table~\ref{tab:6}, we find the reason that yields
such a phenomenon is that 2-character words appear most often in datasets, with the appearance of 147143, 48717 and 49413 times in training, testing and validation set, respectively. Therefore, in our texts, bigram features can effectively capture the likelihood of 2 characters being a legal word, and they are most beneficial to model performance improvement.

Given the experimental results, we use bigram as additional feature for BiLSTM-CRF based word segmenter.

\begin{table}
\caption{Experimental Results with Different N-gram Features in BiLSTM-CRF.}
\label{tab:5}
\begin{tabular}{llll}
\hline\noalign{\smallskip}
Model + feature        & Precision      & Recall         & $F_{1}$-score             \\
\noalign{\smallskip}\hline\noalign{\smallskip}
BiLSTM-CRF      & 95.81          & 95.97          & 95.89          \\
BiLSTM-CRF + Four-gram             & 96.72          & 96.70          & 96.71        \\
BiLSTM-CRF + Trigram   & 97.19          & 97.44          & 97.32          \\
BiLSTM-CRF + Bigram   & \textbf{97.59} &\textbf{97.80}   & \textbf{97.70}    \\
\noalign{\smallskip}\hline
\end{tabular}
\end{table}

\begin{table}
\caption{Statistics of words whose characters are of different lengths.}
\label{tab:6}
\begin{tabular}{llll}
\hline\noalign{\smallskip}
N-character words        & Training set      & Testing set         & Val set             \\
\noalign{\smallskip}\hline\noalign{\smallskip}
N = 2     & 147143  & 48717         & 49413      \\
N = 3     & 30379   & 10043       & 10385   \\
N = 4     &  8187   & 2707    & 2857       \\
\noalign{\smallskip}\hline
\end{tabular}
\end{table}

\begin{figure}[!ht]
\begin{center}
\includegraphics[width=0.48\textwidth]{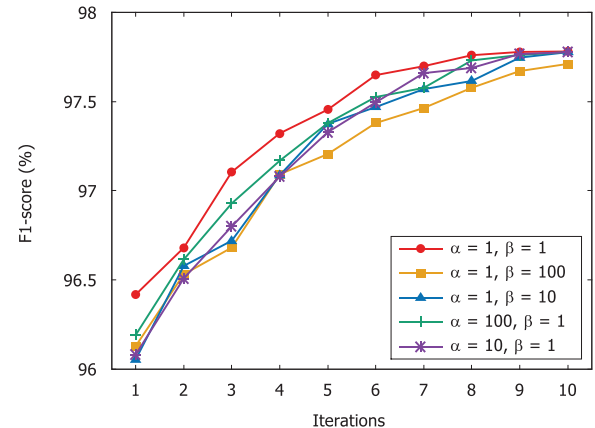}
\caption{Comparisons between different weight coefficients of normalized entropy and loss prediction.}
\label{Fig:weight}
\end{center}
\end{figure}

\subsubsection{Comparisons between Different Weight  Coefficients of Normalized Entropy and Loss Prediction}

To study which part has more influence on the final performance, we conduct an experiment on different weight coefficients of normalized entropy and loss prediction with bigram features. We compare five different groups of parameters in Equation (16).

From the learning curves of Fig.~\ref{Fig:weight}, it is clear that when the weight coefficients $\alpha$ and $\beta$ are all set to 1, the results are better than others in early iterations, and then tend to be uniform, except for the coefficients of 1 and 100.

Furthermore, we find that, when $\alpha$ and $\beta$ are 100 and 1, i.e., we enlarge the effect of loss prediction, the $F_{1}$-scores are higher than the results when $\alpha$ and $\beta$ are 1 and 100. We believe the reason is that loss prediction is task-agnostic as the model is learned from losses regardless of target tasks while normalized entropy is more effective to the task like classification, which is learned to minimize cross-entropy between predictions and labels.

When the weight coefficients $\alpha$ and $\beta$ are all set to 1, respectively, the performance is the best, which shows that combining two parts together can make full use of respective advantages to achieve better results, thus we choose this group of parameters for subsequent experiments.

\begin{figure*}[!ht]
\includegraphics[width=1\textwidth]{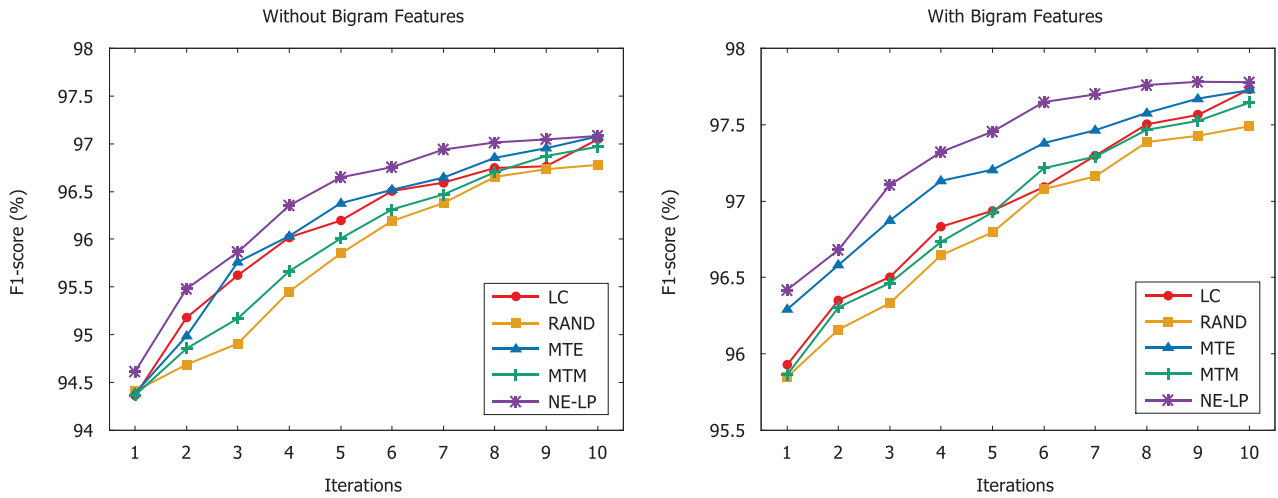}
\caption{Comparisons between different sampling strategies when the ratio of initial labeled set and unlabeled set is 3:7. }
\label{Fig:37}
\end{figure*}

\subsubsection{Comparisons between Different Sampling Strategies}
\label{SubSubSec:different sampling strategies}

In this experiment, we compare the conventional sampling strategies introduced in Section~\ref{Sec:Related Work} with our proposed method \textbf{NE-LP}, as well as the uniformly random baseline (\textbf{RAND}). We evaluate the performance of strategy by its $F_{1}$-score on the testing set. To prove the effectiveness of our proposed method, we conduct our experiments in two configurations: adding additional bigram features and using character-only features. For each iteration, we train 30 epoches with bigram features, which is a good trade-off between speed and performance, while 50 epoches without bigram features to ensure model convergence.

As illustrated in Fig.~\ref{Fig:37}, all sampling strategies perform better than \textbf{RAND} baseline. From the left of Fig.~\ref{Fig:37}, we find that \textbf{LC} and \textbf{MTE} greatly outperform \textbf{MTM} in early rounds while from the right of Fig.~\ref{Fig:37}, we notice that \textbf{MTE} works very effectively with the bigram features, but \textbf{LC} suffers from performance drop. The reason may be that, on the influence of bigram features, \textbf{LC} is not accurate enough to localize the best token to label.

Furthermore, we observe that the $F_{1}$-scores improve greatly when adding bigram features, which again indicates the effectiveness of bigram features.

Regardless of whether to add bigram features, our approach \textbf{NE-LP} shows the best performance for all active learning cycles. The performance gaps between our method \textbf{NE-LP} and entropy-based \textbf{MTE} are obvious since \textbf{NE-LP} not only captures the uncertainty of sequences, but also takes segmentation losses into consideration.

\begin{figure*}[!ht]
\includegraphics[width=1\textwidth]{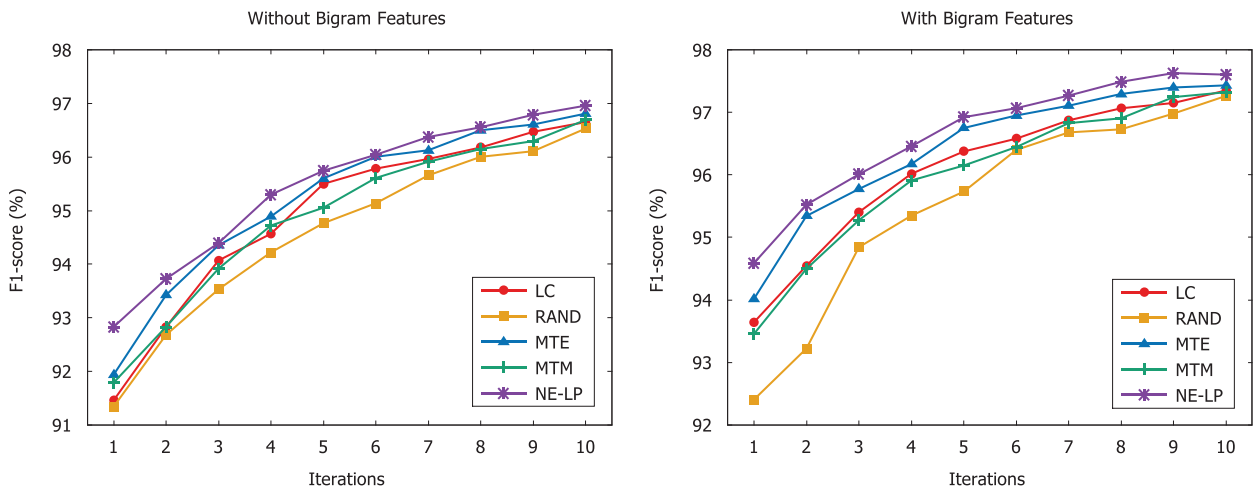}
\caption{Comparisons between different sampling strategies when the ratio of initial labeled set and unlabeled set is 1:9. }
\label{Fig:19}
\end{figure*}

\subsubsection{Comparisons between Different Sampling Strategies with Different Sizes of Initial labeled Set}

Furthermore, we also investigate the effects of different initial labeled set sizes on the final performance. Instead of using the ratio of 3:7, we now divide the training set with the ratio of 1:9 to get the initial labeled set and unlabeled set.

As depicted in Fig.~\ref{Fig:19}, we find that our proposed method \textbf{NE-LP} still outperforms other uncertainty-based sampling strategies at all iterations, which shows that our method can always select informative samples beneficial to current model improvement regardless of the size of initial labeled set.

The performance trends of these sampling strategies are similar to those in Fig.~\ref{Fig:37}. \textbf{NE-LP} shows the best performance, \textbf{MTE} achieves better $F_{1}$-scores than \textbf{LC} and \textbf{MTM} while \textbf{RAND} obtains the lowest results.

However, the performance gaps between \textbf{NE-LP} and \textbf{MTE} are less obvious than Fig.~\ref{Fig:37} since when the ratio is 1:9, losses tend to be smaller than those with the ratio of 3:7. Therefore, in \textbf{NE-LP}, compared to loss prediction, normalized entropy has a greater impact on performance, leading to the phenomenon that the $F_{1}$-score curve of \textbf{NE-LP} is close to \textbf{MTE}. However, despite the small gaps, \textbf{NE-LP} outperforms \textbf{MTE} anyway. Therefore, we still can't ignore the importance of loss prediction since it also plays a role to improve the performance.

\section{Conclusion and Future Work}
\label{Sec:Conclusion}

To relieve the efforts of EHRs annotation, we propose an effective word segmentation method based on active learning with a novel sampling strategy called NE-LP. NE-LP effectively utilizes the output of a joint model and combines normalized entropy with self-attention based loss prediction. Compared to the widely-used and mainstream uncertainty-based sampling methods, our sampling strategy selects samples from statistical perspective and deep learning level. In addition, to capture coherence between characters, we further add bigram features to the joint model. Based on EHRs collected from the Shuguang Hospital Affiliated to Shanghai University of Traditional Chinese Medicine, we evaluate our method on CWS. Compared to conventional uncertainty-based sampling strategies, NE-LP achieves best performance, which proves the effectiveness of our method to a certain extent.

As possible research directions, we plan to employ other highly performant pre-trained neural networks, such as Bert and GPT for EHRs segmentation. Then, considering the characteristics of CWS task and model, we believe that our method can also be applied to other tasks, such as NER and relation extraction.

\begin{acknowledgements}
We kindly thank Ju Gao from Shuguang Hospital Affiliated to Shanghai University of Traditional Chinese Medicine for providing us clinical datasets, and Ping He from Shanghai Hospital Development Center for her help. This work was supported by the National Natural Science Foundation of China (No. 61903144) and the National Key R\&D Program of China for ``Precision medical research" (No. 2018YFC0910550).
\end{acknowledgements}

\section*{Compliance with ethical standards}
\textbf{Conflict of interest} No conflict of interest exits in the submission of this manuscript.

\bibliographystyle{spmpsci}
\bibliography{ReferencesBibfiles}

\begin{thebibliography}{10}
\providecommand{\url}[1]{{#1}}
\providecommand{\urlprefix}{URL }
\expandafter\ifx\csname urlstyle\endcsname\relax
  \providecommand{\doi}[1]{DOI~\discretionary{}{}{}#1}\else
  \providecommand{\doi}{DOI~\discretionary{}{}{}\begingroup
  \urlstyle{rm}\Url}\fi

\bibitem{angluin1988queries}
Angluin, D.: Queries and concept learning.
\newblock Machine Learning \textbf{2}(4), 319--342 (1988)

\bibitem{bengio1994learning}
Bengio, Y., Simard, P., Frasconi, P., et~al.: Learning long-term dependencies
  with gradient descent is difficult.
\newblock IEEE Transactions on Neural Networks \textbf{5}(2), 157--166 (1994)

\bibitem{chen2015long}
Chen, X., Qiu, X., Zhu, C., Liu, P., Huang, X.: Long short-term memory neural
  networks for {Chinese} word segmentation.
\newblock In: Proceedings of the 2015 Conference on Empirical Methods in
  Natural Language Processing, pp. 1197--1206 (2015)

\bibitem{cheng2020sentiment}
Cheng, K., Yue, Y., Song, Z.: Sentiment classification based on part-of-speech
  and self-attention mechanism.
\newblock IEEE Access \textbf{8}, 16387--16396 (2020)

\bibitem{collobert2011natural}
Collobert, R., Weston, J., Bottou, L., Karlen, M., Kavukcuoglu, K., Kuksa, P.:
  Natural language processing (almost) from scratch.
\newblock Journal of Machine Learning Research \textbf{12}(Aug), 2493--2537
  (2011)

\bibitem{eddy1998profile}
Eddy, S.R.: Profile hidden markov models.
\newblock Bioinformatics (Oxford, England) \textbf{14}(9), 755--763 (1998)

\bibitem{gan2019investigating}
Gan, L., Zhang, Y.: Investigating self-attention network for {Chinese} word
  segmentation.
\newblock arXiv preprint arXiv:1907.11512  (2019)

\bibitem{gesulga2017barriers}
Gesulga, J.M., Berjame, A., Moquiala, K.S., Galido, A.: Barriers to electronic
  health record system implementation and information systems resources: A
  structured review.
\newblock Procedia Computer Science \textbf{124}, 544--551 (2017)

\bibitem{gilad2006query}
Gilad-Bachrach, R., Navot, A., Tishby, N.: Query by committee made real.
\newblock In: Advances in neural information processing systems, pp. 443--450
  (2006)

\bibitem{goldberg2014word2vec}
Goldberg, Y., Levy, O.: {Word2Vec} explained: deriving mikolov et al.'s
  negative-sampling word-embedding method.
\newblock arXiv preprint arXiv:1402.3722  (2014)

\bibitem{houlsby2011bayesian}
Houlsby, N., Husz{\'a}r, F., Ghahramani, Z., Lengyel, M.: Bayesian active
  learning for classification and preference learning.
\newblock arXiv preprint arXiv:1112.5745  (2011)

\bibitem{Hu2018Active}
Hu, P., Lipton, Z.C., Anandkumar, A., Ramanan, D.: Active learning with partial
  feedback.
\newblock In: International Conference on Learning Representations (2018)

\bibitem{huang2018low}
Huang, H., Wang, H., Jin, D.: A low-cost named entity recognition research
  based on active learning.
\newblock Scientific Programming \textbf{2018}, 1--10 (2018)

\bibitem{jiang2019seq}
Jiang, W., Tang, Y.: A seq-to-seq transformer premised temporal convolutional
  network for {Chinese} word segmentation.
\newblock arXiv preprint arXiv:1905.08454  (2019)

\bibitem{la2020research}
La~Su, Y., Liu, W.: Research on the lstm mongolian and chinese machine
  translation based on morpheme encoding.
\newblock Neural Computing and Applications \textbf{32}(1), 41--49 (2020)

\bibitem{lafferty2001conditional}
Lafferty, J.D., McCallum, A., Pereira, F.C.: Conditional random fields:
  Probabilistic models for segmenting and labeling sequence data.
\newblock In: Proceedings of the Eighteenth International Conference on Machine
  Learning, pp. 282--289 (2001)

\bibitem{lei2018effective}
Lei, L., Zhou, Y., Zhai, J., Zhang, L., Fang, Z., He, P., Gao, J.: An effective
  patient representation learning for time-series prediction tasks based on
  ehrs.
\newblock In: IEEE International Conference on Bioinformatics and Biomedicine,
  pp. 885--892. IEEE (2018)

\bibitem{lewis1994sequential}
Lewis, D.D., Gale, W.A.: A sequential algorithm for training text classifiers.
\newblock In: Proceedings of the 17th Annual International Conference on
  Research and Development in Information Retrieval, pp. 3--12. Springer (1994)

\bibitem{li2012active}
Li, S., Zhou, G., Huang, C.R.: Active learning for {Chinese} word segmentation.
\newblock In: Proceedings of International Conference on Computational
  Linguistics 2012: Posters, pp. 683--692 (2012)

\bibitem{liu2019neural}
Liu, J., Wu, F., Wu, C., Huang, Y., Xie, X.: Neural chinese word segmentation
  with dictionary.
\newblock Neurocomputing \textbf{338}, 46--54 (2019)

\bibitem{liu2020ltp}
Liu, M., Tu, Z., Wang, Z., Xu, X.: {LTP}: A new active learning strategy for
  {Bert-CRF} based named entity recognition.
\newblock arXiv preprint arXiv:2001.02524  (2020)

\bibitem{ma2018state}
Ma, J., Ganchev, K., Weiss, D.: State-of-the-art {Chinese} word segmentation
  with bi-lstms.
\newblock In: Proceedings of the 2018 Conference on Empirical Methods in
  Natural Language Processing, pp. 4902--4908 (2018)

\bibitem{marcheggiani2014experimental}
Marcheggiani, D., Artieres, T.: An experimental comparison of active learning
  strategies for partially labeled sequences.
\newblock In: Proceedings of the 2014 Conference on Empirical Methods in
  Natural Language Processing, pp. 898--906 (2014)

\bibitem{peng2004chinese}
Peng, F., Feng, F., McCallum, A.: Chinese segmentation and new word detection
  using conditional random fields.
\newblock In: Proceedings of the 20th international conference on Computational
  Linguistics, pp. 562--568. Association for Computational Linguistics (2004)

\bibitem{shao2019domain}
Shao, D., Zheng, N., Yang, Z., Chen, Z., Xiang, Y., Xian, Y., Yu, Z.:
  Domain-specific {Chinese} word segmentation based on bi-directional
  long-short term memory model.
\newblock IEEE Access \textbf{7}, 12993--13002 (2019)

\bibitem{song-etal-2010-active}
Song, H., Yao, T., Kit, C., Cai, D.: Active learning based corpus annotation.
\newblock In: {CIPS}-{SIGHAN} Joint Conference on {C}hinese Language Processing
  (2010)

\bibitem{srivastava2014dropout}
Srivastava, N., Hinton, G., Krizhevsky, A., Sutskever, I., Salakhutdinov, R.:
  Dropout: a simple way to prevent neural networks from overfitting.
\newblock Journal of Machine Learning Research \textbf{15}(1), 1929--1958
  (2014)

\bibitem{sun2020attention}
Sun, D., Yaqot, A., Qiu, J., Rauchhaupt, L., Jumar, U., Wu, H.: Attention-based
  deep convolutional neural network for spectral efficiency optimization in
  mimo systems.
\newblock Neural Computing and Applications  (2020)

\bibitem{tang2020recognizing}
Tang, P., Yang, P., Shi, Y., Zhou, Y., Lin, F., Wang, Y.: Recognizing {Chinese}
  judicial named entity using {BiLSTM-CRF}.
\newblock arXiv preprint arXiv:2006.00464  (2020)

\bibitem{tang2018combining}
Tang, X., Du, B., Huang, J., Wang, Z., Zhang, L.: On combining active and
  transfer learning for medical data classification.
\newblock Institution of Engineering and Technology Computer Vision
  \textbf{13}(2), 194--205 (2018)

\bibitem{vaswani2017attention}
Vaswani, A., Shazeer, N., Parmar, N., Uszkoreit, J., Jones, L., Gomez, A.N.,
  Kaiser, {\L}., Polosukhin, I.: Attention is all you need.
\newblock In: Advances in Neural Information Processing Systems, pp. 5998--6008
  (2017)

\bibitem{wang2017convolutional}
Wang, C., Xu, B.: Convolutional neural network with word embeddings for
  {Chinese} word segmentation.
\newblock arXiv preprint arXiv:1711.04411  (2017)

\bibitem{wang2019incorporating}
Wang, Q., Zhou, Y., Ruan, T., Gao, D., Xia, Y., He, P.: Incorporating
  dictionaries into deep neural networks for the {Chinese} clinical named
  entity recognition.
\newblock Journal of biomedical informatics \textbf{92}, 103--133 (2019)

\bibitem{xing2018adaptive}
Xing, J., Zhu, K., Zhang, S.: Adaptive multi-task transfer learning for
  {Chinese} word segmentation in medical text.
\newblock In: Proceedings of the 27th International Conference on Computational
  Linguistics, pp. 3619--3630 (2018)

\bibitem{xue2003chinese}
Xue, N., Shen, L.: Chinese word segmentation as lmr tagging.
\newblock In: Proceedings of the second SIGHAN workshop on Chinese language
  processing-Volume 17, pp. 176--179. Association for Computational Linguistics
  (2003)

\bibitem{yan2017active}
Yan, Q., Wang, L., Li, S., Liu, H., Zhou, G.: Active learning for {Chinese}
  word segmentation on judgements.
\newblock In: National CCF Conference on Natural Language Processing and
  Chinese Computing, pp. 839--848. Springer (2017)

\bibitem{yang2014overview}
Yang, J., Yu, Q., Guan, Y., Jiang, Z.: An overview of research on electronic
  medical record oriented named entity recognition and entity relation
  extraction.
\newblock Acta Automatica Sinica \textbf{40}(8), 1537--1562 (2014)

\bibitem{yoo2019learning}
Yoo, D., Kweon, I.S.: Learning loss for active learning.
\newblock In: Proceedings of the IEEE Conference on Computer Vision and Pattern
  Recognition, pp. 93--102 (2019)

\bibitem{zhao2006effective}
Zhao, H., Huang, C.N., Li, M., Lu, B.L.: Effective tag set selection in
  {Chinese} word segmentation via conditional random field modeling.
\newblock In: Proceedings of the 20th Pacific Asia Conference on Language,
  Information and Computation, pp. 87--94 (2006)

\bibitem{zheng2013deep}
Zheng, X., Chen, H., Xu, T.: Deep learning for {Chinese} word segmentation and
  pos tagging.
\newblock In: Proceedings of the 2013 Conference on Empirical Methods in
  Natural Language Processing, pp. 647--657 (2013)

\end{thebibliography}
\end{document}